\documentclass{egpubl}
\usepackage{sca2024}

 \SpecialIssuePaper         

\CGFStandardLicense

\usepackage[T1]{fontenc}
\usepackage{dfadobe}  

\biberVersion
\BibtexOrBiblatex
\usepackage[backend=biber,bibstyle=EG,citestyle=alphabetic,backref=true]{biblatex} 
\electronicVersion
\PrintedOrElectronic
\ifpdf \usepackage[pdftex]{graphicx} \pdfcompresslevel=9
\else \usepackage[dvips]{graphicx} \fi

\usepackage{egweblnk} 

\usepackage{todonotes}
\usepackage{xspace}
\usepackage[noabbrev, capitalise]{cleveref}
\usepackage{xcolor}
\usepackage{makecell}
\usepackage[]{algorithm2e}
\usepackage{multirow}
\usepackage{amsfonts}

\usepackage{color}
\definecolor{bcolor}{rgb}     {1.0,0,0}

\title{Learning to Move Like Professional Counter-Strike Players \vspace{-1em}}

\author[D. Durst et al.]
{\parbox{\textwidth}{\centering 
         D. Durst$^{1}$\orcid{0000-0002-4960-0336}
         ~
         F. Xie$^{2}$\orcid{0000-0001-9366-820X} 
         ~
         V. Sarukkai$^{1}$\orcid{0009-0006-9809-9994}
         ~
         B. Shacklett$^{1}$\orcid{0000-0002-2894-1158}
         ~
         I. Frosio$^{3}$\orcid{0000-0002-7230-4287}
         ~
         C. Tessler$^{3}$\orcid{0000-0001-6447-9864}
         ~
         J. Kim$^{3}$\orcid{0000-0001-7385-7732}
         ~
         C. Taylor$^{2}$\orcid{0009-0008-1940-364X}
         ~
         G. Bernstein$^{4}$\orcid{0000-0002-3016-1169}
         \\
         S. Choudhury$^{5}$\orcid{0000-0003-2762-8888}
         ~
         P. Hanrahan$^{1}$\orcid{0000-0002-3474-9752}
         ~
         K. Fatahalian$^{1}$\orcid{0000-0001-8754-0429}
         }
         \\
{\parbox{\textwidth}{\centering 
         $^1$Stanford University
         ~
         $^2$Activision Blizzard
         ~
         $^3$NVIDIA
         ~
         $^4$University of Washington
         ~
         $^5$Cornell University
       }
\vspace{-1.5em}
}
}

\begin{document}

\newcommand{\stanford}{Stanford University}
\newcommand{\cornell}{Cornell University}
\newcommand{\washington}{University of Washington}
\newcommand{\nvidia}{NVIDIA}
\newcommand{\activision}{Activision Blizzard}

\newcommand{\red}[1]{\textcolor{red}{#1}}
\newcommand{\numtestrounds}{1430\;}

\newcommand{\learnedbot}{\textsc{MLMove}\xspace}
\newcommand{\handcraftedbot}{\textsc{RuleMove}\xspace}
\newcommand{\defaultbot}{\textsc{GameBot}\xspace}
\newcommand{\humanbot}{\textsc{Human}\xspace}
\newcommand{\noattention}{\textsc{NoAttn}\xspace}
\newcommand{\posbombonly}{\textsc{PosBomb}\xspace}
\newcommand{\history}{\textsc{History}\xspace} 
\newcommand{\csgo}{CS:GO\xspace}
\newcommand{\dust}{de\_dust2\xspace}
\newcommand{\csknow}{\textsc{CSKnow}\xspace}
\newcommand{\emd}{earth mover's distance}
\newcommand{\motion}{CSMove\xspace}
\newcommand{\images}{images}

\colorlet{yellowish}{green!10!orange}
\colorlet{greenish}{green!70!orange}
\colorlet{blueish}{blue!70}
\colorlet{purpleish}{purple!70}
\colorlet{reddish}{red!70}
\definecolor{greenstage}{RGB}{12,156,19}
\definecolor{purplestage}{RGB}{163,22,180}
\definecolor{yellowstage}{RGB}{191,191,14}

\newif\ifenableimages
\enableimagestrue

\teaser{
\ifenableimages
\includegraphics[width=\textwidth]{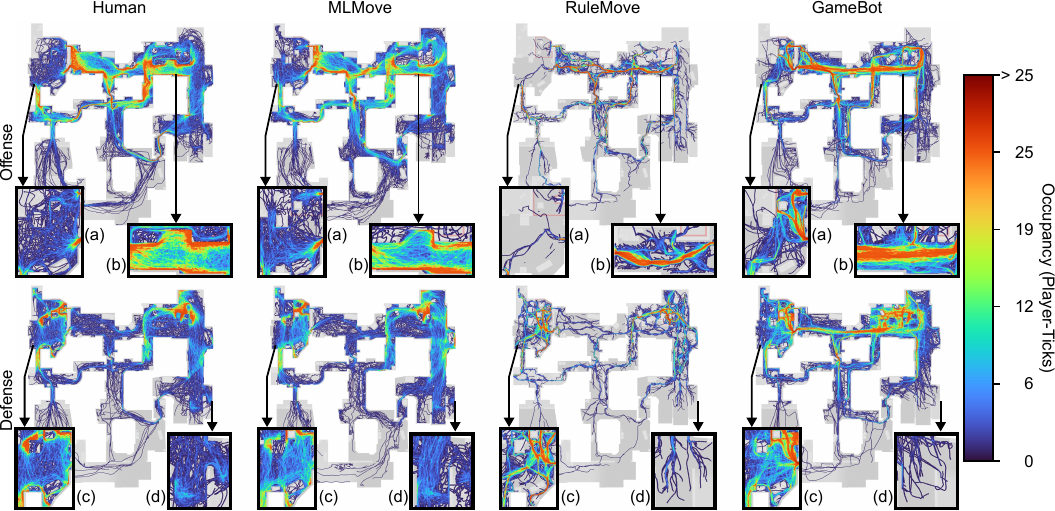}
\fi
\caption{We introduce \learnedbot, a bot for playing \csgo Retakes that features a movement controller trained on logs from 123 hours of professional human play. The controller generates movement actions for both teams of players in 0.5~ms (amortized per-step cost) on a single CPU core. The figures plot the fraction of time players spend in different regions of the map, aggregated over \numtestrounds rounds of play. The distribution of the \learnedbot bots playing against themselves (second column) mimics the overall distribution of human play (\humanbot, first column).
A well-engineered rule-based bot (\handcraftedbot) and the bots currently shipping in \csgo (\defaultbot) do not replicate the human movement distribution.
\label{fig:overall-distribution}
}
}

\maketitle

\begin{abstract}

In multiplayer, first-person shooter games like Counter-Strike: Global Offensive (\csgo), coordinated movement is a critical component of high-level strategic play. 
However, the complexity of team coordination and the variety of conditions present in popular game maps make it impractical to author hand-crafted movement policies for every scenario.
We show that it is possible to take a data-driven approach to creating human-like movement controllers for \csgo.
We curate a team movement dataset comprising 123 hours of professional game play traces, and use this dataset to train a transformer-based movement model that generates human-like team movement for all players in a ``Retakes'' round of the game.  Importantly, the movement prediction model is efficient. Performing inference for all players takes less than 0.5~ms per game step (amortized cost) on a single CPU core, making it plausible for use in commercial games today. Human evaluators assess that our model behaves more like humans than both commercially-available bots and procedural movement controllers scripted by experts (16\% to 59\% higher by TrueSkill rating of ``human-like'').
Using experiments involving in-game bot vs. bot self-play, we demonstrate that our model performs simple forms of teamwork, makes fewer common movement mistakes, and yields movement distributions, player lifetimes, and kill locations similar to those observed in professional \csgo match play.


\end{abstract}

\section{Introduction}

Competitive, multiplayer first-person shooters (FPS) are extraordinarily popular. Multiple titles have tens of millions of users every month~\cite{poole2023cod, danastasio2023valorant}. 
\csgo is one of the seminal titles in the genre, with millions of players each day~\cite{csgo_player_count}.
AI agents (``bots'') that can effectively imitate human players have the potential to improve human players' experiences by serving as training partners for new players and teammates for experienced players when their friends are not available~\cite{gdc_modl2022imitation}.

It is challenging to create a human-like bot for a complex, team-based FPS game like \csgo. 
Existing approaches based on hand-crafted behavior rules or learned models struggle to generate realistic, coordinated player movement due to:
\begin{enumerate}
    \item \textbf{Complexity of human movement.} Hand-crafted, rule-based bots remain the prevalent practice in modern multiplayer FPS games. However, it is not tenable to encode rules spanning the massive number state combinations of $10$ players in a complex $3$D world. As a result, hand-crafted bots lack realism because they fail to react appropriately to a diverse set of game situations.
    \item \textbf{Matching human movement distributions}. While reinforcement learning (RL) approaches have been shown to produce highly-skilled (even superhuman) behavior~\cite{deepmind_jaderberg2019human, openai_five_berner2019dota, silver2018general, lample2017playing}, it is difficult to craft reward functions that yield policies that ``move like humans''. As a result, RL bots fail to serve as good human proxies.
    \item \textbf{Compute efficiency.} Imitation learning (IL) approaches can produce policies that replicate human behavior recorded in player logs, and this approach has been deployed for player controllers in turn-based games~\cite{meta2022human}. However, these games' run-time performance requirements are orders of magnitude (100$\times$) lower than that of a real-time FPS game. 
    Most commercial FPS games require AI controller logic to use only a small fraction of the total per-frame CPU budget~\cite{aaa_ai_budget} (limiting execution to a few ms on a single CPU core~\cite{valorant_compute_budget_2020, gdc_lind2020performance}).  
    Unfortunately, recent work using IL to create FPS bots~\cite{pearce2022counter} requires orders of magnitude (800$\times$) more compute than this limit.
\end{enumerate}

In this paper we present the first compute efficient, data-driven method for creating bots that move like human players in the FPS game Counter-Strike: Global Offensive (\csgo). Our bots, which include a small transformer-based model~\cite{waymo_ngiam2022scene} trained using imitation learning, move like experienced human team players, execute well-within the AI budget of commercial FPS games, and are simple and fast to train. 
Specifically, our work makes the following contributions:

\textbf{(1)} \textbf{Efficient transformer-based movement controller.} 
We present the first compute-efficient, transformer-based model specialized for controlling movement in \csgo, called \learnedbot. Our model focuses on playing one map (\dust) and one game mode (Retakes). Once trained through standard supervised learning, \learnedbot produces human-like movement actions in response to evolving game dynamics. Our movement model's amortized runtime cost for controlling two teams of bots in a \csgo match is just under 0.5~ms per game step on a single CPU core (8~ms inference every 16~game steps), meeting commercial game servers' performance requirement. Human evaluators assess that our model is more human-like than both commercially-available bots and expert-crafted rule-based movement bots by 16\% to 59\% (according to a TrueSkill rating) in the user study. 

\textbf{(2)} \textbf{Pro-player \csgo movement dataset curation system.}  We create a system for the curation of a 123-hour dataset of \csgo game play called \csknow.  This is the first large scale dataset curated for learning team-based movement in a popular FPS game featuring professional players. 

\textbf{(3)} \textbf{Quantitative positioning metrics for assessing human-like behavior.} Our goal is to produce realistic bot movement at both short-term and longer-term (full round) time scales. We define novel quantitative metrics computed on rounds of bot vs. bot self-play that assess how well a bot's movement emulates human players' team-based positioning.
We demonstrate these metrics correlate with the human evaluators' assessment of human-like game play.

We refer the reader to \url{https://mlmove.github.io} for the open-source system including the trained transformer-based movement model, the rule-based execution module, the \csknow dataset curation system, and the complete Python evaluation code.

\section{Related Work}
\label{sec:related_work}

Human-like agent navigation is an important component of multiple applications including robotics, autonomous driving, visual effects, and games. For example, crowd simulation in games and visual effects endeavor to generate trajectories for hundreds or thousands of simple agents with much less inter-agent interactions than FPS games~\cite{reynolds1987Boids,wolinski2017Warpdriver,panayiotou2022configurableCrowd}; while embodied agent motion planning research for robot navigation requires orders of magnitude more compute resources than FPS games to interact with a real physical world observed through cameras~\cite{zheng2024towards,huang2023visual,endres2012evaluation}.

Our work addresses the challenge of human-like motion control for groups of autonomous agents in the context of FPS games where the agents have to perform a wide range of movements (walk, run, jump) in a dynamic environment under extreme run-time performance constraints.  The most closely related work to ours fall into three categories: 
\begin{enumerate}
    \item hand-crafted, rule-based controllers where developers must manually encode all of the controller's behaviors
    \item RL-based controllers where developers specify a reward function that the controller maximizes
    \item IL-based controllers where developers specify a set of human examples that the controller imitates
\end{enumerate}

\noindent \textbf{Rule-Based Multi-Agent Movement Controllers.} 
A common abstraction for organizing rule-based controllers is behavior trees~\cite{gdc_isla2005bt}. 
However, rule-based approaches struggle to generate human-like behavior in more complex environments, as demonstrated by Huang et al.'s complex hierarchy for coordinating movement of pedestrians through doorways~\cite{huang2018doorway}.  As a result, human evaluators find state-of-the-art rule-based bots for FPS games often make unhuman-like movement decisions. See our final evaluation section for more details.

\noindent \textbf{RL Multi-Agent Movement Controllers.}
RL agents can generate superhuman behavior in complex strategy games like Dota 2 and Go by maximizing a reward function~\cite{silver2017mastering,openai_five_berner2019dota}.
RL can also be used to train agents to win in FPS games like Doom and Quake~\cite{lample2017playing, deepmind_jaderberg2019human}.
However, these types of RL agents are not trained to act like humans because they are trained to maximize a reward function for winning.
Humans may struggle to collaborate with the RL agents whose actions do not match human expectations~\cite{meta2022human}.
In contrast, we use an IL-based approach to create a bot that moves like humans.

\noindent \textbf{IL Multi-Agent Movement Controllers.}
When trained on large, diverse training sets of human play, IL-based controllers can generate human-like movement for a wide range of situations.
Scene Transformer~\cite{waymo_ngiam2022scene} trained a transformer for predicting multiple pedestrians' and  cars' trajectories on different roads over a five-second time horizon.  Scene Transformer leverages the transformer's attention mechanism to learn relationships between cars, pedestrians, and road geometry.
MotionLM~\cite{waymo2023motionlm} demonstrated that a decoder-only transformer architecture can increase accuracy, since the decoder enforces causal relationships between earlier and later time steps.
The models used in Scene Transformer and MotionLM cannot be directly applied to motion control for FPS games, because their compute cost is multiple orders of magnitude too high (their target use case is around five seconds per query).
Adapt~\cite{aydemir2023adapt}, a compute-optimized movement model based on the Scene Transformer, runs in $11$~ms when highly optimized for a Tesla T4 GPU. Its compute cost is still at least two orders of magnitude greater than the AI budget of FPS games~\cite{nvidia_teslat4}. In contrast, our transformer model, designed for learning human-like movement in team-based FPS games, requires two orders of magnitude less compute than Adapt without any hardware specific optimization.  

Existing IL bots are also too computationally expensive for commercial FPS games. \cite{pearce2022counter} trained a model that controls all game behavior (not just movement) of a single \csgo bot using rendered images as input.
This pixels-to-action approach (similar to \cite{kanervisto2020benchmarking, guss2019minerl}) requires a GPU for every agent, approximately three order of magnitude higher compute than commercial FPS games' AI performance constraints. 
Additionally, \cite{pearce2022counter} do not generate coordinated team behavior because they train on data from, and test their bots in, a game mode where players typically practice low-level mechanics without the need for intra-team coordination.

Research on hybrid RL and IL training procedures enable reward-based approaches that also generate human-like behavior.
GREIL is an RL-based crowd control policy trained with a reward function based on similarity to human examples~\cite{charalambous2023greil}.
Cicero is a bot trained with piKL, which regularizes the reward function with an IL policy to prevent drastic deviation from human behavior.
Cicero is designed for Diplomacy, a turn-based strategy game where action frequency is 100 times slower than in an FPS~\cite{meta2022human}. 
We are not aware of a hybrid RL/IL approach for human-like bots in an FPS game.

\section{Problem Formulation: A Bot for \csgo Retakes}
\label{sec:overview}

\noindent \textbf{Game Context.}
\csgo is a multiplayer FPS involving two teams competing for control over a map.
To focus on player movement, we concentrate our attention on a popular \csgo practice mode known as ``Retakes'' and on a single map, \dust. Even though FPS games like \csgo can have many maps, maps are designed to have similar room and path layout characteristics that are known to enable interesting game play; and expert players tend to hone their strategy by playing on the same map over and over. For these reasons, we chose to focus our study on the extremely popular \dust.

The rules of \csgo ``Retakes'' are the same regardless of map choice.
In each round, a bomb is planted in one of two pre-determined regions  known as bombsites A and B. 
(We provide an illustrated example of a standard game map with annotated bombsites regions in Section 4 of the Supplemental Material.)
The bomb will explode in 40 seconds unless it is defused.
The goal of one team, who we call the defense, is to defend the bomb until it explodes.
At most $3$ players are on defense.
The goal of the other team, who we call the offense, is to defuse the bomb before it explodes. 
At most $4$ players are on offense.
One defense player must start at the bomb location while all other players can start at any location on the map.
Members of the two teams can eliminate each other using several weapons and grenades.
Without losing generality, we restrict all players to the same weapon type and preclude the use of any grenades.

\noindent \textbf{State.} The game state at time $t$ consists of player states $q_{i,t} \in Q_t$ as well as global game state that consists of the map state $map_t$ and key events $e_{i, t} \in E_t$ like players shooting or being eliminated.
Time~$t$ is tracked inside each round of \csgo using game ticks. For the rest of this paper, we use game ticks (steps) and time $t$ interchangeably.
We use $\mathbb{B}$ to represent $ \{\mathrm{True}, \mathrm{False}\}$ and $\mathbb{Z}$ to represent the set of all integers.
\begin{enumerate}
    \item Each player's state $q_{i,t} = [p_{i,t}, v_{i,t}, u_i, l_{i,t}, vd_{i,t}, h_{i,t}, r_{i,t}]$ consists of position $p_{i,t} \in \mathbb{R}^3$, velocity $v_{i,t} \in \mathbb{R}^3$, team $u_{i} \in \{\mathrm{Offense}, \mathrm{Defense}\}$, alive status $l_{i,t} \in \mathbb{B}$, view direction $vd_{i,t} \in \mathbb{R}^2$, health $h_{i,t} \in \mathbb{Z}$, and armor $r_{i,t} \in \mathbb{Z}$. 
    \item Map state $map_t = [b_t, x_t]$ consists of the target bombsite $b \in \{\mathrm{A},\mathrm{B}\}$ and seconds left until the bomb explodes $x_t \in \mathbb{R}$.  
    \item Each game event $e_{i,t} = [src_{i,t}, tgt_{i,t}, y_{i,t}]$ consists of source player id $src_{i,t} \in \mathbb{Z}$, optional target player id $tgt_{i,t} \in \mathbb{Z}$, and type $y_{i,t} \in \{\mathrm{shoot}, \mathrm{hurt}, \mathrm{elimination}\}$.
\end{enumerate}

\noindent \textbf{Actions.} A player's action at time step $t$ $a_{i,t} = [m_{i,t}, du_{i,t}, f_{i,t}]$ consists of movement command $m_{i,t} \in \mathbb{Z}$ specifying which direction to move, how fast, and whether to jump or not; aim command a.k.a view direction update command $du_{i,t} \in \mathbb{R}^2$; and fire command $fc_{i,t} \in \mathbb{B}$.

\begin{figure*}[!ht]
    \centering
    \ifenableimages
    \includegraphics[width=\textwidth]{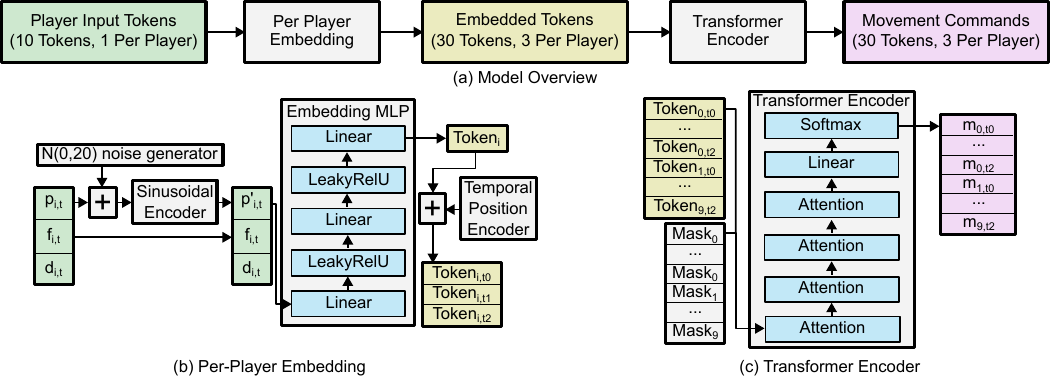}
    \fi
    \caption{\textbf{The learned movement model.} (a) shows an overview of the two stages: (1) the per-player embedding stage converts the input tokens into embedded tokens, and (2) the transformer encoder uses the embedded tokens to predict the movement commands. (b) shows the per-player embedding stage that converts each player input token to three embedded tokens using a three layer MLP. (c) shows the transformer encoder that uses the embedded tokens and the associated masks to predict each player's movement command probabilities. 
    }
    \label{fig:model-architecture}
    \vspace{-2em}
\end{figure*}

\noindent \textbf{Objective.} 
Create a \csgo bot that plays like an expert human in a team play setting. We note that the objective of playing like an expert human in a team setting is not the same as playing to win.

Expert human players utilize complex strategies that require spatial and temporal coordination to defeat their opponents. While these types of strategies are difficult to emulate for rule-based agents, we observe that most team play strategies revolve around positioning players in optimal locations for defeating the other team. The other in-game actions, like aiming and firing, can be effectively predicted using rules once player positions are determined.

Therefore, in this paper, we design and build an IL-based movement model capable of generating movement commands $m_{i,t}$ for all players such that our bots move and position themselves like humans; then we use rule-based execution  to emulate professional players' aiming and firing behavior.

\section{\learnedbot: A Learned \csgo Bot}
\label{sec:model}

In this section, we present the algorithm and system design for our human-like \csgo Retakes bot \learnedbot.
First, we present our transformer-based movement model, then we present how we integrate our movement model with a rule-based command executor to create \learnedbot.

\subsection{Learned Movement Model}
\label{subsec:model-arch}

The main challenges of building a movement model that emulates expert human players in a team-based FPS game are the conflicting goals of accurately predicting the distribution of complex human actions and extremely efficient compute usage.

In an FPS game, human players not only have a complex action space, but also demonstrate complex inter-player interaction and coordination that are quite challenging to model in a rule-based system. However, recent work in transformer models show how to imitate the \textit{effect} of complex human decisions and interactions without modeling the \textit{intermediate} steps (or decisions) that led to the final actions.

The architecture of our movement controller (\cref{fig:model-architecture}) is inspired by Scene Transformer~\cite{waymo_ngiam2022scene}, one of many~\cite{aydemir2023adapt, waymo2023motionlm, yuan2021agentformer} transformer-based multi-agent motion prediction system for pedestrians and autonomous vehicles. The Scene Transformer encodes the state of each agent as input tokens and leverages attention to learn the relationships between all the agents. 
Like Scene Transformer, our movement controller can also benefit from the transformer architecture's ability to capture rich player interactions with the attention mechanism, process players' state in any order due to the permutation invariance of input tokens, and handle eliminated players with attention  masking~\cite{vaswani2017attention}.  

However, Scene Transformer's query latency and compute resources were orders of magnitude higher than what was acceptable for FPS game adoption.  Our key insight was to leverage the significant differences between the target applications of FPS games and autonomous vehicles to make architecture and system design choices to create a movement model that: (a) is able to emulate the effect of complex human team play strategy and interactions in an FPS game, and (b) can be executed within the strict compute constraint required by FPS game servers. We highlight two of these differences below.

First, multi-agent motion prediction systems for autonomous vehicles must support changing road geometry (dynamic road graphs)  as road conditions can change as vehicles travel from one part of the real world to another. On the other hand, professional players for FPS games tend to play and compete on the same game map for \textit{years}, and map geometry and layout are static throughout the game, so it's perfectly reasonable to design movement models that are trained for one map.  This design choice allowed us to reduce the number of input tokens significantly, and as a result, the complexity of our attention layers, without impacting our model's applicability for our targeted use case. Note that each attention layer's complexity is proportional to the square of the number of input tokens~\cite{vaswani2017attention}.
To support multiple maps, we can pre-train our model for each map we want to support and make them available to our \learnedbot bot. (We would of course also have to curate a training set for each map as well, just like the training dataset for Scene Transformer includes data spanning multiple map regions.)

Second, Scene Transformer predicts motion trajectories for up to five seconds, a standard latency measure for pedestrian motion prediction.  To capture the causal dependency from later to earlier predicted positions across the same five second time interval, Scene Transformer uses an encoder-decoder architecture.  For an FPS game where human players can make navigation decisions at 125~ms intervals (due to average keyboard latency),  the motion controller only needs to predict movement trajectories that hold for 125~ms (the time interval at which it is invoked), an order of magnitude lower than five seconds.  For our use case, we can use a simpler and more efficient encoder architecture where all output tokens are computed in parallel. Additionally, the shorter prediction time horizon also translates to fewer input and output tokens and reduced complexity in our attention layers.

These application driven design choices enabled us to create a model that can predict human-like movement decisions for two teams of \csgo players (10 total players) within $8$~ms per query on a single CPU core.

\noindent \textbf{Model Input.} 
Our movement model's input is a sequence of $10$ tokens, each token describing a player's current state.
\csgo logs contain up to 10 players at any time, up to five on each team. This is a broader range of players than in Retakes. We train our model on $10$ input tokens to enable it to generalize to a wider range of situations.
The feature vector of each token is $[p_{i,t}, f_{i,t}, d_{i,t}]$, where $f_{i,t} = [l_{i,t}, u_{i}, b_t, x_t]$ and $d_{i,t}$ is a set of derived features that approximate information not contained in the game logs like visibility and team communication about strategy.
Each token starts with the player's position $p_{i,t}$, alive status $l_{i_t}$, and team association $u_i$. We also include in each token the global map states of bomb location $b_t$ and remaining time for bomb explosion $x_t$, information known to all players.  
We define the derived features in Section 1.1 of the Supplemental Material.
We found that the derived features can aid attention in limited situations.

\noindent \textbf{Model Output.} 
Our movement model's output is a sequence of tokens, each token describing a player's movement command: which direction to move, how fast, and whether to jump.
To capture the multi-modal and stochastic nature of player movement, we represent a movement command as a discrete probability distribution with 97 options.  
Each option corresponds to a combination of one of 16 angular directions, three different movement speeds, two jumping vs not jumping states; plus a separate option for standing still.
Movement commands aren't recorded in \csgo logs.
We use heuristics to infer the movement commands from position/velocity information in the logs~\cite{pearce2022counter}. 
We found discretizing direction uniformly into 16 absolute angles is sufficient to navigate game map details like thin ledges.

\noindent \textbf{Model Architecture.} 
The full architecture of the movement model is depicted in \cref{fig:model-architecture}(a). Each input player token is converted by an embedding network to an embedded token of dimension matching that of the transformer's attention layers; then each sequence of 10 embedded tokens corresponding to the states of 10 players are processed by the transformer to yield the movement command probabilities for the 10 players. 

We use a learned embedding (\cref{fig:model-architecture}(b)) to convert input player tokens into vectors of dimension $256$. Our embedding network consists of three linear layers, with LeakyReLU activations in between the linear layers. 
Our transformer encoder (\cref{fig:model-architecture}(c)), consists of four identical single-head self attention layers of dimension $256$.  Like~\cite{vaswani2017attention}, we use a learned linear transformation and softmax to convert the outputs of our attention network to predicted probability of the output tokens (the player movement commands in our case).

To support eliminated players, we use transformer's masking feature.
A transformer's attention layer computes the attention (connection) between all token pairs in the input sequence except for those that are masked out.
So we set $mask(i, t) = 1$ for each token of an eliminated player ($l_{i,t} = \mathrm{false}$), to remove attention between that player and all other players.
Also, we restrict the loss computation to only use $P(m_i)$ for players that are alive.
Together with attention masking, this ensures eliminated players have no impact on our model's movement predictions for live players.

To learn temporally coherent motions, our model outputs predictions not only for the immediate next action (0~ms into the future) but also for actions at 125~ms and 250~ms from the current time. This is achieved by replicating each player's embedded token for time $t$ three times, and summing each player's embedded token with the positional encoding of the three timestamps, to create distinctive embedded tokens for current time $t$, $t+125~\mathrm{ms}$, and $t+250~\mathrm{ms}$. Like \cite{vaswani2017attention, waymo_ngiam2022scene}, we use sinusoidal positional encoding for the player's in-game map position and for the three temporal positions represented as timestamps.

A well known problem in imitation learning is the inertia problem, where models trained on sequences are biased to repeating recent actions, since this type of repeating ``what I did last'' behavior tends to dominate the dataset~\cite{de2019causal, codevilla2019exploring, seo2023regularized}. This can lead to the failure to learn important movements like velocity change or (intentionally acted) ``erratic'' movements in combat,  because they are both rare (low probability) events in the training dataset.
We address the inertia problem using a simple solution that improves our model's prediction accuracy and efficiency: our model input consists only of ``current'' player states. 
The ablation in \cref{subsec:ablation} shows our solution's effectiveness, as adding prior input states leads to less human-like map occupancy and kill location distributions.

\noindent \textbf{Model Training}\label{subsec: movement training}
We train the movement model using standard supervised learning where we minimize the cross-entropy loss between the probability distributions of the predicted movement command and the ground truth movement commands in the dataset.

We train using the \csknow dataset (described \cref{sec:data}) and perform an 80/20 train-test split: 5655508 train data points (98 hours at 16~Hz) and 1429953 test data points (25 hours at 16~Hz).
Since there is a strong correlation between data points in the same round, we assign all data points in each round to the same subset.
Once grouped into train/test subsets, we randomly sort data points irrespective of their round.
We use the same train/test split for all training runs.
To improve the model's ability to generalize, we add random Gaussian noise with mean~0 and variance~20 \csgo units (less than a player's width of 32 units) to the player positions (see \cref{fig:model-architecture}(b)).

We train for 20 epochs with a batch size of 1024, an initial learning rate of 4e-5 controlled by the Adam optimizer with default configuration ($\beta$s = (0.9, 0.999), eps = 1e-08, and weight decay = 0).
Training takes 1.5 hours on a single computer with a Intel i7-12700K CPU, 128 GB of RAM, and an NVIDIA RTX 4090.

\subsection{Integrating Movement Model into a \csgo Bot}\label{subsec: inference}
The resulting trained model predicts the movement for all players efficiently enough to be deployed in a commercial FPS game server. Specifically, the memory requirement for our model's 5.4M parameters is 21 MB; and the inference latency (time it takes to predict the movement of all players) of our trained movement model deployed in C++ using LibTorch and TorchScript~\cite{torchscript} is 8~ms with an IQR of 0.6~ms on a single core of a Intel Xeon 8375C CPU.

\begin{figure}
    \centering
    \ifenableimages
    \includegraphics[width=\columnwidth]{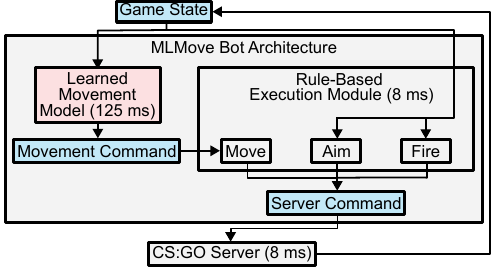}
    \fi
    \caption{\textbf{\learnedbot Architecture:} \learnedbot uses the learned movement model to generate movement commands, then it uses a rule-based execution module to convert these commands into keyboard actions and also to generate aiming and firing commands.
    \csgo server executes all player commands and sends the updated game state back to the bot.}
    \label{fig:bot-diagram}
    \vspace{-1em}
\end{figure}

\begin{figure*}
    \centering
    \ifenableimages
    \includegraphics[width=\textwidth]{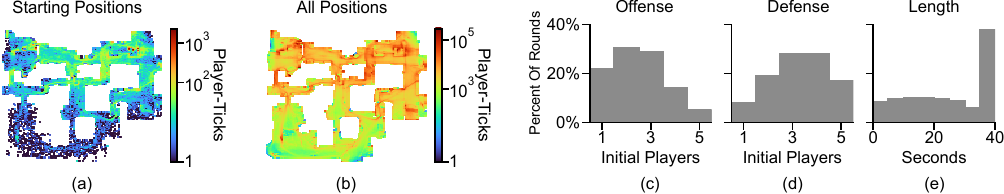}
    \fi
    \caption{CSKnow is a diverse dataset of 123 hours of professional \csgo play. (a) Density of player positions at the start of each round. Players start in a wide range of positions. (b) Density of player positions throughout the entire round. Players visit all areas of the map. (c)-(d) Rounds start with different numbers of offense and defense players, and can end almost immediately or last until the explosion (e). Note:
    (a)-(b) graphs are log scale, (c)-(e) graphs are linear scale. 
    }
    \label{fig:dataset-coverage}
    \vspace{-2em}
\end{figure*}

We use a modular approach shown in \cref{fig:bot-diagram} to integrate our learned movement model into a full \csgo bot \learnedbot. 
The input to \learnedbot is the current game state and output of the \learnedbot is a sequence of game server commands that can be produced and sent to the server by a regular human player.

The core system of \learnedbot consists of our learned movement model and the rule-based execution module. Every 125~ms (16 game ticks), \learnedbot requests the learned movement model to predict the movement commands for all players using the current game player state as input. 
The bot caches and reuses the predicted movement commands for the subsequent 125~ms. The rule-based execution module is executed every game tick to emulate human mouse movement latency used for aiming. It converts movement commands at current time $t$ made by our learned model into human players' keyboard navigation commands. The movement commands are only updated once every 125~ms to emulate human keyboard press latency; therefore, the amortized compute cost for our learned movement model for each game tick (frame) is 0.5~ms.

Our rule-based execution module also generates aiming and firing commands based on the current game state and player positions; the rule-based execution module sends all the ``machine generated'' game commands to the server, which will execute the commands and update the game state for the next frame. 

For human and bot mixed play, the game server just replaces bot generated commands with the corresponding human player's commands. 
As addressed in \cref{sec:evaluation}, we primarily test games consisting only of bots. 
However, the server allows any mixture of human and bot players for a maximum of ten players.

\noindent \textbf{Aiming and firing} 
We use standard techniques to handle aiming and firing.
If no enemy is visible, the aiming module uses a probabilistic occupancy map to pick a target where enemies are likely to appear, emulating human-like predictive aim~\cite{occupancy_moravec1988sensor, occupancy_isla2013third}.
If at least one enemy is visible, the module selects one target and tracks them until they are no longer visible.
The aim module generates a smooth trajectory of view direction updates using a semi-implicit Euler method~\cite{bargteil2020introduction}.
The fire execution module emits fire commands when the crosshair aligns with an enemy's axis-aligned bounding box. 
A distance-based lookup table controls the fire command frequency, shooting shorter and more controllable sequences at farther enemies that are harder to hit.

For specific implementation details, see Section 3 of the Supplemental Material.

\section{CSKnow Dataset Curation System}
\label{sec:data}

There is a scarcity of open datasets for learning movement control for FPS games. Prior \csgo datasets focused on long-term outcomes like win probability, so they captured game state at too low frequency for evaluating movement commands at every 125~ms.  For example, ESTA contains professional game play with data points every 500~ms~\cite{data_xenopoulous2022esta}, and PureSkill.GG contains amateur game play with no guarantees on data capture frequency or even if some data were dropped~\cite{data_pureskill2024}.

We present \csknow, the first dataset for learning team-based \csgo movement featuring professional players.
The dataset contains 123~hours of play sampled at 16~Hz. The data comes from over 17K rounds, features 2292 unique players, 513K shots, and 29K eliminations.
See Section 1 of the Supplemental Material for the subset of game state extracted in \csknow.

We created a system to curate the 123~hour dataset from logs of 1156 hours played on the \dust map by professional players between April 2021 and November 2022.
We downloaded the logs from HLTV~\cite{hltvwebsite}.
The logs contain game play from the complete \csgo game mode, not just the Retakes practice one. 
Unlike the Retakes mode, the complete game mode requires five on each team at the start of each round and involves an earlier stage where teams compete to plant the bomb.
We filter the data in \csknow to game ticks when the bomb has been planted and at least one player is alive on both teams, a super-set of Retakes.
This filter ensures our dataset is focused enough to be representative of Retakes mode play style while still broad enough to cover a diverse range of game play.

\cref{fig:dataset-coverage} shows that \csknow covers a diverse range of play situations: players start in a wide variety of starting positions, and over the course of play move into all locations on the map.  
Since bomb plants in a full game occur in the middle of \csgo rounds, the number of players that are alive on each team at the time of bomb plant varies significantly in the data set.
\section{Evaluation}
\label{sec:evaluation}

The primary goal of our work is to produce human-like movement for an FPS game.
To evaluate how well we achieved this goal, we first conducted a small-scale user study (inspired by BotPrize 2010~\cite{hingston2010new, hingston2009turing}) where human evaluators rank movement in videos of games played by humans and bots, and an exploratory study where humans play with and against the bots.
Then, we performed a large-scale quantitative comparison on the distributions of movement trajectories and key outcomes from bot vs. bot self-play relative to those from professional human play. 
Through this combination of small-scale human ranking and large-scale quantitative analysis of outcome distributions, we present the first comprehensive evaluation of human-like team-based movement for multiplayer FPS bots. 

\begin{figure}
    \centering
    \ifenableimages
    \includegraphics[width=\columnwidth]{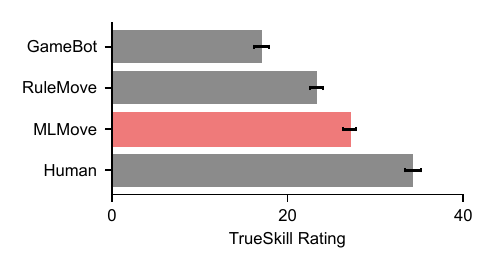}
    \fi
    \caption{Human evaluators consistently rated \learnedbot's behavior as more human than \handcraftedbot and \defaultbot.}
    \label{fig:userstudy}
    \vspace{-2em}
\end{figure}

\subsection{Experiment Conditions.} 
\label{subsec:conditions}
We compare four different player configurations:
\begin{itemize}
\item \humanbot. Replay of the actual human data, taken from the \csknow dataset. 
\item \learnedbot. Our bot with a learned movement controller, as described in \cref{sec:model}. 
\item \handcraftedbot. A bot with a rule-based movement controller implemented by the authors. The bot uses the same rule-based aim and firing controllers as \learnedbot. This bot was developed over several months by a skilled \csgo player and should be considered a strong baseline for \csgo bot design. Section 3 of the Supplemental Material provides further detail on this bot's logic. 
\item \defaultbot. The bots currently deployed in the commercial \csgo game. Since it is third-party commercial software, the implementation details of this bot are unknown. It differs from \learnedbot and \handcraftedbot in movement, aiming, and firing.
\end{itemize}

There are \numtestrounds rounds in the \csknow test dataset that meet Retakes conditions. 
Our results analyze play from full Retakes rounds where all players are controlled using the same player configuration, such as \learnedbot vs \learnedbot with no \humanbot in the game.
In the user study, we randomly sample $8$ rounds across a range of initial conditions and record $32$ videos, one for each combination of player configuration and round. 
For each round, participants viewed all four videos in a random order without labels identifying the player configuration.
We used \csgo to generate videos of game play, rendered from a birds-eye camera position and angle that best enabled analysis of team-based movement. For evaluator clarity, we used the ``x-ray vision'' rendering mode so evaluators can see players behind walls. 
The videos have a median length of 17 seconds and an IQR length of 17 seconds. 
We provide all 32 videos as well as the specific prompts of the study in the Supplemental Material. 
In the quantitative self-play experiments, we ran each player configuration through five iterations of all \numtestrounds rounds in order to account for randomness in game play.

\subsection{Human Assessment}
\label{sec:userstudy}
To assess the realism of bot motion, we conducted a within-subjects study where we asked human evaluators to watch \csgo game play videos depicting both human and bot play~\cite{park2023generative}. 
For each of the eight rounds described in \cref{subsec:conditions}, participants were asked to rank player configurations based on how well player movement matched their ``expectation of how humans would move in that situation.''  

\noindent \textbf{Evaluators.}
We recruited fifteen evaluators with \csgo experience ranging from novice (never having played) to expert. Five of them achieved a rank of ``Global Elite'', the highest \csgo player rating; and four had a rank of ``Supreme Master First Class``, the second highest.

\noindent \textbf{Quantitative Ranking Results.}
Our study produces 120 rankings of the player configurations.
Each ranking is an ordering of the four player configurations' similarities to expected human behavior in one initial condition according to one evaluator.
To enable comparison between player configurations across all rankings, we use the TrueSkill rating~\cite{herbrich2006trueskill} to aggregate the data into a single rating for each player.
TrueSkill is a generalization of the Elo~\cite{glickman1995comprehensive} rating system to multiplayer environments.
In our work, a higher ranking means that a player configuration better matches the evaluators' expectations of human behavior.

In \cref{fig:userstudy} we plot the mean and standard deviation of the TrueSkill rating value for each player configuration. Unsurprisingly, \humanbot achieves the best rating, whereas \learnedbot generates motion that matches evaluator expectations for human movement significantly more frequently than the other bots. The results also suggest that \handcraftedbot is a strong baseline, since it achieves a higher rating than \defaultbot, which is in commercial use today.  The results are statistically significant according to a Kruskal-Wallis test (H=333, p<1e-5) and Dunn post-hoc tests (all p<1e-5).

\noindent \textbf{Qualitative User Feedback}
In addition to ranking the player configurations, subjects were also asked to explain their decisions. 
Expert subjects report that \learnedbot players demonstrated coarse-grained teamwork like ``trading'': killing an enemy while that enemy was distracted engaging someone else. 
Trading is a result of team-based movement, as two teammates must be in the right places at the right time to setup and take advantage of an enemy's momentary weakness.
However, they also reported observing teamwork-related \learnedbot mistakes, such as being overly aggressive when trading, overly passive when supporting an attacking teammate, and lacking temporal coherence by rechecking previously cleared areas or jittering forwards and backwards. 
Experts complemented \humanbot on their skilled collaborative movement, and criticized \handcraftedbot and \defaultbot as bot-like.
\handcraftedbot was too rigid, and \defaultbot made illogical decisions.

\subsection{Quantitative Self-Play Experiments Analysis}
\label{exp: traj matching}

\begin{table}[]
    \centering
    \caption{Median $\pm$ IQR earth mover's distance (EMD) between map occupancy distributions (\cref{sec:eval:positions}), player kill location distributions (\cref{sec:eval:outcomes}), round lifetime distributions, and shots per kill distributions created from bot self-play and from real human data. In all metrics, self-play using \learnedbot yields distributions that are more similar to \humanbot than \handcraftedbot. We attribute the increased distance between lifetime distributions from \learnedbot and human play to an increased number of long lifetime trajectories caused by instances of passive \learnedbot play (see \cref{sec:eval:outcomes}).}
    
    \begin{tabular}{ r | r@{\hspace{0.2em}}l r@{\hspace{0em}}r r@{\hspace{0em}}r } 
    EMD Type & \multicolumn{2}{r}{\learnedbot} & \multicolumn{2}{r}{\handcraftedbot} & \multicolumn{2}{r}{\defaultbot} \\
        \hline
        Map Occupancy & \textbf{8.2 $\pm$} & \textbf{0.5} & 14.7 $\pm$ & 1.7 & 15.2 $\pm$ & 0.3 \\
        \hline
        Kill Locations & \textbf{6.7 $\pm$} & \textbf{0.1} & 15.4 $\pm$ & 0.7 & 16.4 $\pm$ & 0.7 \\
        \hline
        Lifetimes & 4.9 $\pm$ & 0.4 & 7.8 $\pm$ & 0.0 & \textbf{1.1 $\pm$} & \textbf{0.0} \\
        \hline
        Shots Per Kill & \textbf{2.1 $\pm$} & \textbf{0.1} & 5.6 $\pm$ & 0.0 & 4.9 $\pm$ & 0.2
    \end{tabular}
    \label{tab:emd}
    \vspace{-1em}
\end{table}

\begin{figure}
    \centering
    \ifenableimages
    \includegraphics[width=\columnwidth]{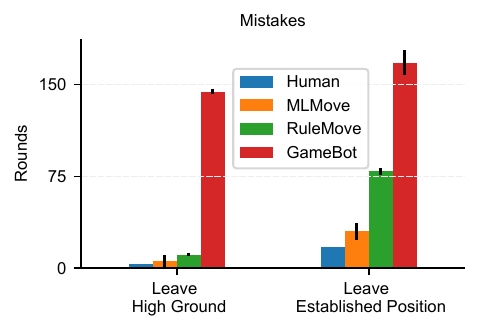}
    \fi
    \caption{Median and IQR counts of rounds where at least one defensive player makes one of two common positioning mistakes (leaving high ground and leaving an established defensive position). \learnedbot makes these mistakes far less often than \defaultbot and \handcraftedbot. }
    \label{fig:mistakes}
    \vspace{-1em}
\end{figure}

Beyond the user study, we provide a quantitative evaluation of the four player configurations by analyzing the statistics of full rounds of in-game self-play.
Our metrics cover the key properties of movement: map coverage, utilizing expert strategies that avoid low-skill mistakes, and ensuring that movement yields key outcomes.
The metrics covering mistakes and teamwork rely on a key insight: expert human language is the way to measure planning.
We quantify teamwork and mistakes by utilizing expert labels for map regions, and then quantifying how humans navigate these regions in space and time.
The use of expert terminology to formalize plans sets the stage for future work on foundation models to plan human movement using expert language.
In Section 5.5 of the Supplemental Material, we quantitatively evaluate the ability of the learned movement model to reproduce humans' sequences of actions.

We perform our quantitative analysis on \numtestrounds rounds (550 minutes) of \learnedbot; this is $\sim 5-16 \times$ larger than the quantitative analysis on prior \csgo bots by \cite{pearce2022counter} and on BotPrize bots by \cite{gamez2012neurally}, who inspired our use of Earth Mover's Distance and position-based metrics. 
For summary metrics, we report the median and IQR of the five round iterations discussed in \cref{subsec:conditions}.
For distribution visualizations, we report results from the first iteration for each bot in order to compare distributions with the same numbers of rounds as \humanbot.

\subsubsection{Distribution of Player Positions}
\label{sec:eval:positions}

\cref{fig:overall-distribution} shows the distribution of player positions across the first iteration of \numtestrounds rounds. Each pixel counts the game ticks when an offense or defense player occupies that location of the map. Overall, the distributions of the \learnedbot positions appear more similar to that of \humanbot players than any of the other bot player configurations, for both the offense and defense teams. The first row in Table~\ref{tab:emd} shows that, when measured using earth mover's distance~\cite{kolouri2017optimal}, the \learnedbot occupancy distribution (computed over both the offense and defense teams) is 1.8$\times$ and 1.9$\times$ more similar to that of \humanbot play than \handcraftedbot and \defaultbot respectively. We provide details of how we compute EMD in Section 5.3 of the Supplemental Material.

\cref{fig:overall-distribution} also shows that \learnedbot players exhibit skilled movement characteristics such as positioning themselves to remain out of enemy sight lines. For example,  \learnedbot{s} stay near the walls on offense (inset (b)), and close to objects used for cover on defense (inset (c)), whereas the other bots traverse dangerous areas out in the open. \learnedbot players also demonstrate a greater diversity of behaviors than \handcraftedbot, where each different behavior must be scripted. Insets (a) and (d) highlight examples where \learnedbot echos the diversity of real-world play, but \handcraftedbot follows a limited set of predefined paths.

We also observe situations where \learnedbot produces movement that differs from the human trajectories in important ways. For example, inspection of \cref{fig:overall-distribution} suggests that the model fails to turn corners as sharply as \humanbot. 
\humanbot's inset (b) has more paths near bent walls than \learnedbot's because the humans can turn more sharply to follow the bends. 
We've found that \learnedbot's turning radius limitation is particularly detrimental in an area of the map that requires navigating consecutive tight turns followed by stairs.

\subsubsection{Avoiding Common Mistakes}
\label{sec:eval:mistakes}

A first trait of ``nonhuman'' bot behavior is ``a lack of common sense'', which can be measured by the number of ``common'' mistakes.  We consider two mistakes: (a) leaving high ground, or (b) giving up on an established defensive position.
To characterize these mistakes, we identify specific combinations of players' positions within regions of the map indicating a defensive advantage on a game tick.
For each such scenario, we compute whether the defensive players' regions in the next game tick indicate that they gave up their advantage. 
We measure the number of rounds with at least one mistake.
As shown in \cref{fig:mistakes}, \learnedbot's mistake rate is close to that of human players, and significantly smaller than those of the other bots.

\subsubsection{Teamwork}
\label{sec:eval:teamwork}

\begin{table}
    \caption{Median $\pm$ IQR absolute percentage error (when counting instances of flanking and spreading configurations that arise out of teamwork) of bot players compared to human play data. \learnedbot more closely matches the human distribution of these multiplayer teamwork behaviors.} 
    \begin{center}
        \begin{tabular}{ r | r@{\hspace{0em}}r r@{\hspace{0em}}r } 
            & \multicolumn{2}{r}{Offense} &  \multicolumn{2}{r}{Defense} \\
            \hline
            \learnedbot & \textbf{27}\% \textbf{$\pm$} & \textbf{22}\% & \textbf{13}\% \textbf{$\pm$} & \textbf{14}\% \\
            \hline
            \handcraftedbot & 55\% $\pm$ & 29\% & 42\% $\pm$ & 134\% \\
            \hline
            \defaultbot & 58\% $\pm$ & 23\% & 87\% $\pm$ & 203\%
        \end{tabular}
    \end{center}
    \label{tab:teamwork}
\end{table}

\begin{figure*}[!ht]
    \centering    
    \ifenableimages
    \includegraphics[width=\textwidth]{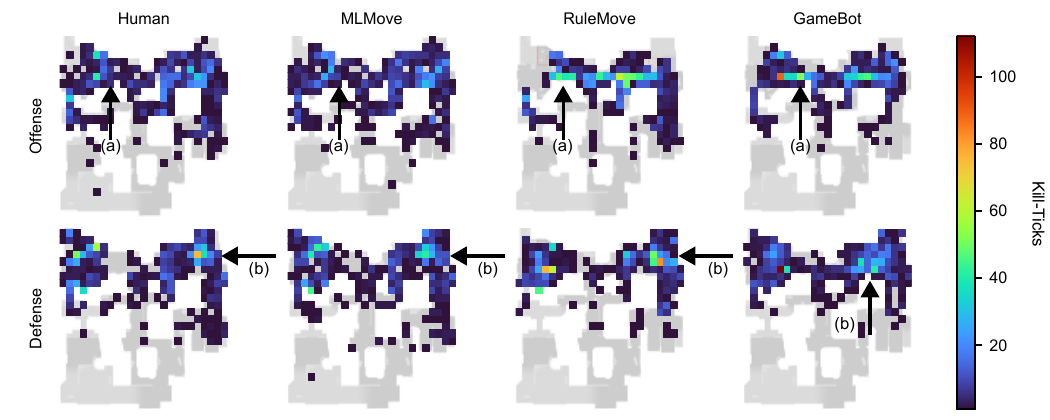}
    \fi
    \caption{Visualization of the number of kills scored at each location on the map (position of shooter). (a) \learnedbot and humans avoid getting into combat in open areas, while \handcraftedbot and \defaultbot frequently record kills from the center of the map, indicating bad positioning. (b) \learnedbot, \handcraftedbot, and humans all score a high number of kills from positions of cover in the center of bombsite A, while kill locations of \defaultbot are more spread out.}
    \label{fig:kill-positions}
    \vspace{-1em}
\end{figure*}

We analyze the self-play rounds for instances of common forms of teamwork. Specifically we focus on \emph{offense flanking}, where multiple players on offense approach the defense from different directions to catch the defenders off guard. We also count instances of \emph{defense spreading}, a tactic where defense players carefully distance themselves so that each player can cover a different potential attack direction, while being close enough to quickly reconverge on the most important actual attack direction.

We identify five unique two-player flanking configurations (involving different combinations of attack directions) and six unique three-player spreading configurations (covering different attack directions), and count the number of rounds where these configurations are observed. 
We define each configuration as a combination of the map regions occupied simultaneously by players on the same team.
We compute the number of rounds with at least one occurrence of each configuration.

\cref{tab:teamwork} shows that \learnedbot not only exhibits all five flanking and all six spreading strategies, but it also employs these strategies with a frequency more similar to human play than the non-learned bots. 
The median absolute percent error between human and \learnedbot flanking counts is 27\%, far less than 55\% and 58\% for \handcraftedbot and \defaultbot respectively. The median absolute percent error between human and \learnedbot spreading counts is 13\%, far less than the 42\% and 87\% for \handcraftedbot and \defaultbot respectively. 
See Section 5.2 of the Supplemental Material for details on the definitions of and results for the individual flanking and spreading configurations.

\subsubsection{Self-Play Outcomes}
\label{sec:eval:outcomes}

Skilled \csgo players move to advantageous positions that increase the likelihood of eliminating enemies without being eliminated.
We hypothesize that if \learnedbot moves similarly to human players, then we will observe similar distributions of where players are located when they score kills, how many shots are taken per enemy kill, and how long players live during rounds.

\noindent \textbf{Kill Locations} \cref{fig:kill-positions} plots the distribution of positions where players score kills (shooter locations), separated into offense and defense teams. Both humans and \learnedbot follow a cover principle when shooting enemies: they tend to shoot more frequently from positions that are protected. In~\cref{fig:kill-positions}(a), both offense humans and \learnedbot avoid combat in the open areas leading to the B bombsite, whereas \handcraftedbot and \defaultbot have poor positioning and engage in these cover-free regions.
In~\cref{fig:kill-positions}(b), defense humans, \learnedbot, and \handcraftedbot (due to map-specific rules) primarily score kills from the center of the A bombsite, where the map contains objects that provide cover. On the other hand, \defaultbot scores kills uniformly around the entire bombsite.
Row~2 of \cref{tab:emd} quantitatively confirms that the \learnedbot's kill location distributions are most similar to human play.

\begin{figure}[]
    \centering
    \ifenableimages
    \includegraphics[width=\columnwidth]{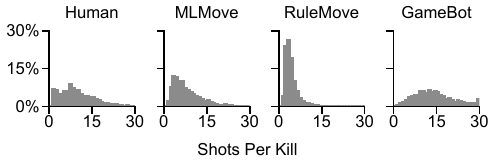}
    \fi
    \caption{In \csgo combat, players attempt to balance conflicting movement goals of staying still (to increase shot accuracy) and unpredictable movement (to avoid fire). \learnedbot reproduces the human distribution of shots per kill. \handcraftedbot is scripted to stop prior to shooting, which leads to higher accuracy shots (fewer shots per kill), but contributes to shorter lifetimes.}
    \label{fig:shots-per-kill}
    \vspace{-1em}
\end{figure}

\noindent \textbf{Shots per kill} We also observe that rounds involving \learnedbot-controlled players demonstrate approximately the same distribution of shots per kill as humans (\cref{fig:shots-per-kill}).
Although it uses the same aiming and firing controller as \learnedbot, \handcraftedbot produces a left-sided distribution, indicating fewer shots per kill.
\handcraftedbot's movement controller tells it to stop moving whenever an enemy becomes visible to increase its shot accuracy, but this behavior is not something all experienced human players would do in practice or in our training dataset.

\begin{figure}[]
    \centering
    \ifenableimages
    \includegraphics[width=\columnwidth]{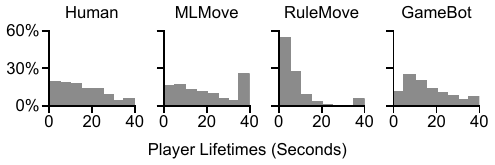}
    \fi
    \caption{The \learnedbot and \defaultbot reproduce the human lifetimes, while \handcraftedbot's tendency to run at the enemy, regardless of the game state, leads to earlier deaths.}
    \label{fig:lifetime}
    \vspace{-2em}
\end{figure}

\noindent \textbf{Player lifetimes} Finally, we observe that rounds involving \learnedbot players exhibit a similarly shaped distribution of player lifetimes as that of human play (\cref{fig:lifetime}).  However, we also observe many more examples of \learnedbot players staying alive for the full 40-second period. 
We believe this is due to a conservative game play strategy present in the \csknow dataset but not in the Retakes test subset. A detailed analysis of this strategy is reported in Sections 1.1 and 5.6 of the Supplemental Material.

\subsubsection{Ablations}
\label{subsec:ablation}
We validate our movement model's design choices using ablations that compare the use of attention and the use of prior states versus without them. 
\cref{tab:ablation-masking} shows results for our movement model (referenced as the default model in this section) in column 2,  our model without attention (\noattention) in column 3, and our model with prior player states added to the input (\history) in column 4. 
Removing attention decreases model accuracy because the model fails to learn relationships between players which affect game play outcomes.
\noattention performs worst on Kill Locations, but also decreases model accuracy on map occupancy and shots per kill.
Adding prior player states causes the inertia problem where players repeat their prior actions rather than responding to the dynamic changes in game states, resulting in worse map occupancy, kill locations, and lifetimes.

All models in \cref{tab:ablation-masking} have a similar inference latency. Our default model has a median inference latency of $6.9$~ms and IQR of $0.6$~ms on one Intel 8375C CPU core, less than $8$~ms. Ablations in Section 5.4 of the Supplemental Material show that increasing the number of attention layers and the size of the MLPs inside each attention layer moderately improves map occupancy similarity to \humanbot distribution while increasing inference latency.

\begin{table}[h]
    \centering
    \caption{Median $\pm$ IQR EMD metrics for the ablated learned movement models. \learnedbot shows our movement model, \noattention shows our movement model with all attention masked out, and \history shows our movement model with prior state added to model input.} 
    \begin{tabular}{ r | r@{\hspace{0.2em}}l r r r} 
    EMD Type & \multicolumn{2}{r}{\learnedbot} & \noattention & \history \\
        \hline
        Map Occupancy & \textbf{8.2 $\pm$} & \textbf{0.5} & 10.3 & 11.8 \\
        \hline
        Kill Locations & \textbf{6.7 $\pm$} & \textbf{0.1} & 8.2 & 7.4 \\
        \hline
        Lifetimes & 4.9 $\pm$ & 0.4 & \textbf{4.6} & 7.7 \\
        \hline
        Shots Per Kill & 2.1 $\pm$ & 0.1 & 2.2 & \textbf{1.2}
        \end{tabular}
    \label{tab:ablation-masking}
    \vspace{-1.5em}
\end{table}

\section{Discussion}
\label{sec:discussion}
We present \learnedbot, the first \csgo bot that uses a learned movement model for generating team-based, human-like movement that satisfies commercial games' performance constraints.

We showed \learnedbot is able to control two full teams of bots with behaviors that match a range of human game play characteristics. Human evaluators ranked \learnedbot as more human than \handcraftedbot and \defaultbot baselines by 16\% to 59\% according to TrueSkill ratings.
We also performed an exploratory user study where experts play with and against the bots.
While the users reported \defaultbot as the least human-like, the study was inconclusive because users reported being too engrossed in the game to evaluate other players' movements in a short, highly controlled experiment.

Our movement model trains in 1.5 hours on a single GPU (attractive for modern game design workflows). The amortized inference cost per game step for our model is less than 0.5~ms on a single CPU core for all players,  making it plausible for commercial game server deployment. 

While our work focused on \csgo, our movement model architecture and training methodology should generalize to other multiplayer FPS games as we mainly leveraged common traits of FPS games in our design (rather than \csgo specific features). However, for each new FPS game, a dataset of size and coverage similar to \csknow is needed to train a transformer-based movement controller similar to ours.  There might be some game specific changes needed to the model input and output to match each game's navigation features. For example, another game might have four instead of three speeds or a few more complex movement modes (like climbing along ledges or ropes).  To create a full bot for a new FPS game that shows human-like team-based movement like \learnedbot, one would also need to build a rule-based execution module similar to the one we used in \learnedbot that can perform aiming and firing controls for the new FPS game, then integrate it with the learned movement model that was customized and trained specifically for this game.  

We also anticipate that our approach to learning human-like movement from data can generalize to other FPS game actions and multiple maps. This would require additions to the input and output tokens; behaviors like firing can be added as parameters of the output tokens, and multiple maps can be supported by adding map geometry encoding as input tokens (i.e., Scene Transformer~\cite{waymo_ngiam2022scene}). 
The main challenges for a more general model with more parameters and tokens would be data collection, model tuning, training complexity, and runtime execution efficiency. Future work can improve performance by using specialized inference engines rather than LibTorch, or creating specialized deployable models (our presented work can be seen as an example) from the general model by reducing non-essential features.

\section{Acknowledgments}
\label{sec:acknowledgments}

Support for this project was provided by Meta, Activision, Andreessen Horowitz, the US ARL under No. W911NF-21-2-0251, the National Science Foundation (NSF) Graduate Research Fellowship Program under Grant No. (DGE-1656518), and a Stanford Graduate Fellowship.
Any opinions, findings, and conclusions or recommendations expressed in this material are those of the author(s) and do not necessarily reflect the views of the sponsors.


\printbibliography     

\end{document}